\documentclass[letterpaper, 10 pt, conference]{ieeeconf}  

\IEEEoverridecommandlockouts                              
\usepackage{amsmath}    			
\usepackage{amssymb}
\usepackage{amsfonts}

\usepackage{graphicx}  				

\usepackage[croatian]{babel}  
\usepackage[utf8]{inputenc}  	
\usepackage[T1]{fontenc}
\usepackage{ae,aecompl}     	

\usepackage{microtype}				

\usepackage{tabularx}
\usepackage{booktabs}
\usepackage{enumerate}				
\usepackage{algorithm2e}			
\usepackage{todonotes}				
\usepackage{dirtree}					
\usepackage{hyperref}					

\usepackage[capitalise]{cleveref}
\usepackage{subcaption}

\usepackage{lmodern}
\usepackage{nccmath}
\usepackage{flushend}
\usepackage{scalerel,stackengine}
\stackMath
\newcommand\reallywidehat[1]{%
	\savestack{\tmpbox}{\stretchto{%
			\scaleto{%
				\scalerel*[\widthof{\ensuremath{#1}}]{\kern.1pt\mathchar"0362\kern.1pt}%
				{\rule{0ex}{\textheight}}
			}{\textheight}%
		}{2.4ex}}%
	\stackon[-6.9pt]{#1}{\tmpbox}%
}
\usepackage{calrsfs}
\DeclareMathAlphabet{\pazocal}{OMS}{zplm}{m}{n}

\graphicspath{{./figures/}}
\usepackage{float}

\begin{document}
\urlstyle{same}
\thispagestyle{empty}
\onecolumn

© 2025 IEEE.  Personal use of this material is permitted.  Permission from IEEE must be obtained for all other uses, in any current or future media, including reprinting/republishing this material for advertising or promotional purposes, creating new collective works, for resale or redistribution to servers or lists, or reuse of any copyrighted component of this work in other works.
\newline

2025 International Conference on Unmanned Aircraft Systems (ICUAS)

DOI: 10.1109/ICUAS65942.2025.11007857

URL: \url{https://ieeexplore.ieee.org/abstract/document/11007857}
\url{}

\twocolumn

\title{\LARGE \bf
Aerial Maritime Vessel Detection and Identification
}
\author{Antonella Barisic Kulas, Frano Petric, Stjepan Bogdan
	\thanks{Authors are with University of Zagreb Faculty of Electrical Engineering and Computing, Unska 3, 10000 Zagreb, Croatia
        {\tt\small (antonella.barisic, frano.petric, stjepan.bogdan) at fer.unizg.hr}}}%

\makeatletter
\newcommand{\removelatexerror}{\let\@latex@error\@gobble}
\newcommand{\mb}[1]{\boldsymbol{#1}}
\newcommand{\norm}[1]{\left\lVert#1\right\rVert}

\makeatother

\markboth{IEEE ICUAS 2025. PREPRINT VERSION.}%
{Barisic Kulas \MakeLowercase{\textit{et al.}}: Aerial Maritime Vessel Detection and Identification}

\maketitle

\thispagestyle{empty}
\pagestyle{empty}

\begin{abstract}

Autonomous maritime surveillance and target vessel identification in environments where Global Navigation Satellite Systems (GNSS) are not available is critical for a number of applications such as search and rescue and threat detection. When the target vessel is only described by visual cues and its last known position is not available, unmanned aerial vehicles (UAVs) must rely solely on on-board vision to scan a large search area under strict computational constraints. To address this challenge, we leverage the YOLOv8 object detection model to detect all vessels in the field of view. We then apply feature matching and hue histogram distance analysis to determine whether any detected vessel corresponds to the target. When found, we localize the target using simple geometric principles. We demonstrate the proposed method in real-world experiments during the MBZIRC2023 competition, integrated into a fully autonomous system with GNSS-denied navigation. We also evaluate the impact of perspective on detection accuracy and localization precision and compare it with the oracle approach.




\end{abstract}

\section{Introduction}
\label{sec:introduction}

Unmanned aerial vehicles (UAVs) have demonstrated exceptional utility across diverse applications, including environmental monitoring \cite{Aucone2023}, infrastructure inspection \cite{Ivanovic2021}, and security operations \cite{Barisic2022mrs}. Within this wide range of applications, the detection of boats using aerial imagery is particularly important and presents a pivotal component in search and rescue (SAR) operations, maritime safety and environmental protection. The agility of drones, combined with their ability to cover vast areas quickly and operate in challenging and remote maritime environments, makes them uniquely suited for efficient and effective maritime surveillance.

\begin{figure}[htb]
    \centering
    \includegraphics[width=\columnwidth]{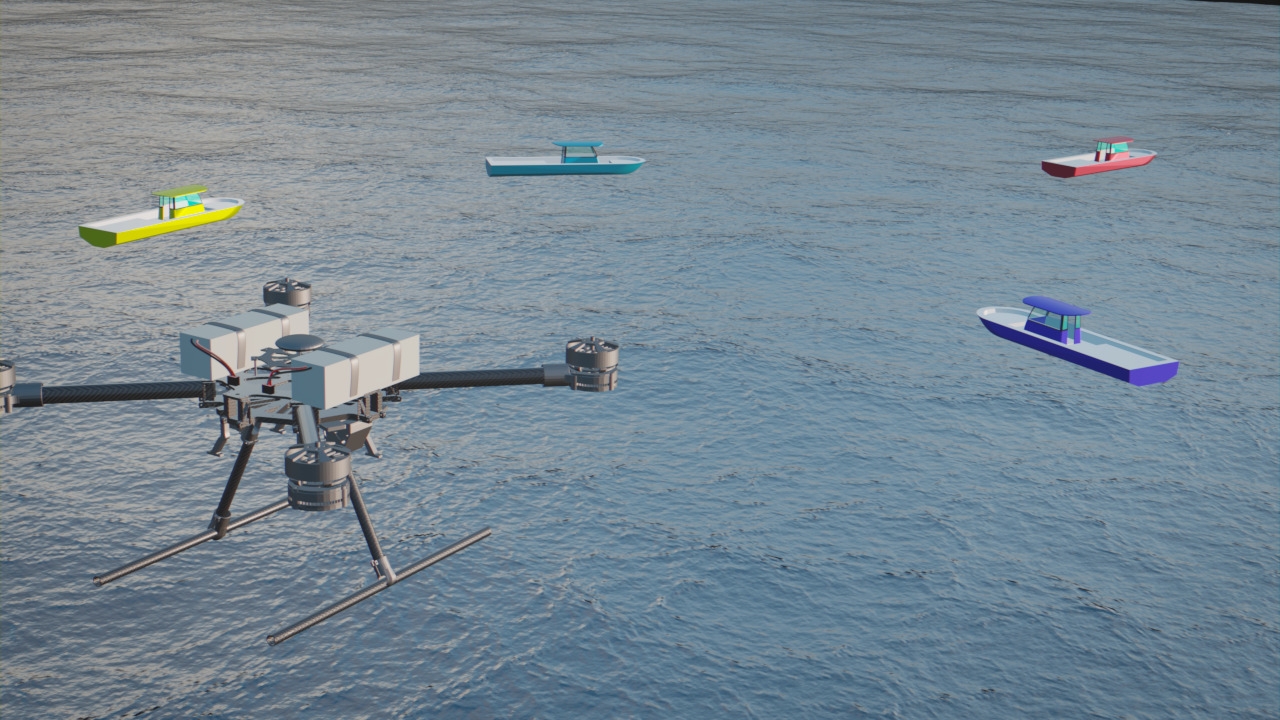}
    \caption{Aerial maritime surveillance part of the MBZIRC 2023 competition. The UAV, flying over large, featureless and GNSS-denied marine area, is tasked with finding, identifying and locating a known vessel among similar vessels.}
    \label{fig:cover}
\end{figure}

Monitoring maritime areas presents significant challenges due to the vast, featureless environments and the small size of vessels relative to the UAV’s field of view, particularly at higher altitudes. These challenges are further exacerbated by dynamic environmental conditions, such as changing lighting, sea reflections, and weather variability, which introduce noise and complexity to aerial imagery \cite{zhao}. To enable timely responses in SAR operations and to combat illegal activities such as smuggling and marine pollution, UAVs require advanced onboard sensing systems that integrate robust detection algorithms and real-time processing capabilities. 

Another barrier to effective UAV deployment in maritime environments is the potential loss or degradation of Global Navigation Satellite System (GNSS) signals. GNSS-denied navigation is essential when satellite signals are jammed, obstructed, or simply unavailable. This necessitates robust onboard localization and sensing solutions that can sustain autonomy even when external positioning data cannot be relied upon.

In this work, we address the challenge of a known target vessel detection through the integration of a neural network-based detection framework with a lightweight identification module that employs background subtraction, object feature extraction, and hue histogram analysis. This methodology demonstrates a successful application in maritime environments, showcasing real-time deployment on a UAV equipped with computationally efficient, low-cost hardware. 

This work is inspired by the MBZIRC 2023 Maritime Challenge (Fig. \ref{fig:cover}), which required teams to develop seamless collaboration between UAVs and USVs for complex navigation and manipulation tasks in GNSS-denied marine environments. The UNIZG-FER team achieved first place in the competition, and in this paper, we present our aerial target vessel identification system and present a comprehensive experimental evaluation.




\section{Related Work}
\label{sec:related_work}

Due to low-cost, availability, greater resolution and ease of deployment, the UAVs have become the most desirable aerial imaging and surveillance platform, surpassing traditional airplane and satellite imaging. This is especially evident in areas of traffic monitoring \cite{du_eccv} or people tracking \cite{stanford_dataset}. With prevalence of such applications, there is also a prevalence of land-based datasets for object detection, recognition and tracking in aerial images. Maritime aerial datasets focusing on ships, and consequently ship detection methods, are relatively scarce, especially taking into account that satellite images are often referred to as "aerial", such as in \cite{airbus}, include only top-down images from larger heights, such as in \cite{kaggle_boat}, or are focused on detection of people in the sea for search and rescue purposes \cite{MOBDrone2021}\cite{varga2022seadronessee}\cite{yoloow}. 

The approach to detection and identification of a known target proposed in this paper relies on both convolutional neural networks (CNNs) and traditional computer vision techniques. CNNs have been the cornerstone of image classification and object detection ever since deployment of AlexNet \cite{krizhevsky2012imagenet}. For ship detection, we deploy YOLOv8s \cite{Jocher_Ultralytics_YOLO_2023} model that was pre-trained on the COCO \cite{coco} dataset, and refine the model using obtained and generated datasets presented in section \ref{sec:datasets}. Following a generic ship detection, we employ feature matching using ORB \cite{Rublee2011} features, which are proven to be significantly faster than the industry standard SIFT \cite{lowe1999object} features. In the final step, we match detections using color histograms, which enables us to fine-tune the identification to different scenarios, giving more weight to feature-matching in closer ranges and emphasizing color match in long range detections dominated by blur and lack of sharpness. Modern feature detection and matching techniques based on deep-learning, such as LightGlue \cite{lightglue} or RoMa \cite{roma}, that are able to match objects from different perspectives, struggle with lack of sharpness of the object at large distances, which, alongside computational complexity, makes them unsuitable for the MBZIRC 2023 scenario. 



\section{MBZIRC 2023 overview}

In this paper, we address the MBZIRC 2023 challenge of UAV-based patrolling over a large maritime region (\(\sim 10\) square kilometers) in response to reports of suspicious activity from satellites, vessels, or similar sources. Prior to deployment, the UAV receives images depicting a specific target boat to be located. The mission must be conducted without GNSS data for navigation or the target’s last known position. In addition, the target boat remains in motion. Therefore, the perception system must identify the target boat quickly and reliably among other boats and work in real time to enable rapid intervention.

\begin{figure*}[htb]
    \centering
    \includegraphics[width=2\columnwidth]{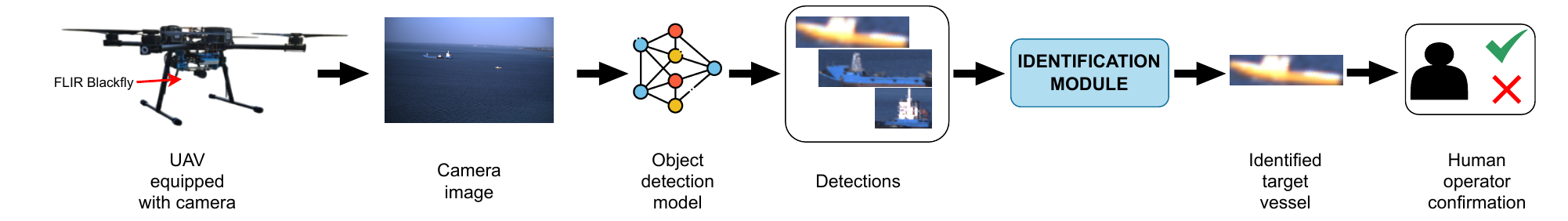}
    \caption{Perception framework}
    \label{fig:framework}
\end{figure*}

Our customised quadcopter UAV is equipped with a front-facing FLIR Blackfly USB3 camera designated for visual maritime surveillance. All onboard processing, including image processing, control algorithms, state estimation and navigation, is performed on an Intel NUC. The UAV uses a down-facing Intel RealSense D435 camera for takeoff and landing on specially designed pads with AR tags, and two Semtech SX1280 LoRa modules for ranging and localization. 

To enable robust operation in GNSS-denied featureless environments in the MBZIRC 2023 competition, we developed a UAV system combining Radio Frequency (RF) ranging via LoRa modules with vision-based techniques for reliable takeoffs and landings. A Kalman filter was employed to effectively fuse noisy RF measurements, ensuring consistent localization, while a dynamic controller/estimator switching mechanism facilitated smooth transitions between multiple localization sources. This system autonomously scanned expansive marine areas, detected and identified target vessels through onboard computer vision, and collaborated with an unmanned surface vehicle (USV) for inspection and intervention tasks—all without dependence on GNSS signals, achieving robust autonomy in challenging maritime scenarios.

The perception framework deployed on the UAV is shown in Fig. \ref{fig:framework}. The images captured by the FLIR camera are processed with the YOLOv8 object detection model, which identifies all vessels present on the sea surface. Each detected vessel is considered as a candidate for further analysis to determine whether it corresponds to the target boat. The identification module evaluates these detections to confirm whether the target vessel has been observed. Once the autonomous system verifies the presence of the target, it transmits the video stream to a human operator for confirmation and further action, reconstructs the position of the target and informs the USV that is tasked with interception of the target. In the following sections, the parts of the proposed solution are described in detail.

\section{Methodology}
\label{sec:target_framework}

\subsection{Datasets}\label{sec:datasets}

The quality and diversity of the data that we train on deep neural networks directly affects their ability to generalize well and to output accurate detections. To that end, we search and collect all boat datasets available. The details of the datasets are presented in the \cref{tab:datasets}. The perspective of the training dataset should represent conditions where object detector will be deployed to ensure best performance. In our search, we found that the availability of online aerial data is less compared to those captured with boat-to-boat perspective. This observation aligns with the expectation that obtaining aerial data is more challenging for logistical reasons. Therefore, to increase data quantity we also use one boat-to-boat dataset. However having in mind that aerial perspective should be dominant in our training data, we collect the most of the data from aerial perspective. The ration of boat-to-boat vs aerial perspective in the real data part of our data is $26:74$. In addition to real data, we also generate two new synthetic datasets to increase diversity of lighting conditions and boat orientations. The MARUSBoats is generated in MARUS simulator, a Unity3D-based simulator specialized for marine robotics, and TargetSynth is generated in Blender using a procedural pipeline detailed in \cite{Barisic2022}. The real-world images of target model are collected in Abu Dhabi during preparation for the competition finals, and are labeled using pseudo-labeling technique with model trained on collected datasets. The ratio of real to synthetic data is $76:24$. Total amount of images used for training is $24858$.

\begin{table}[htb]
  \centering
  \caption{Datasets}
  \begin{tabular}{@{}l|c|c|c@{}}
    \toprule
    Source & Type & Perspective & Img\# \\
    \midrule
    Datasense@CRAS \cite{datasensecras} & real & boat-to-boat & 6511 \\
    Ship Detection \cite{kaggle_boat}   & real &  aerial       & 621  \\
    Aerial Maritime \cite{aerial-maritime_dataset}  & real & aerial & 49\\
    SAR \cite{sar_dataset} & real & aerial & 1127 \\
    SeaDronesSee \cite{varga2022seadronessee} & real & aerial & 8930 \\
    MARUSBoats \cite{marus} & synth & aerial & 5136 \\  
    TargetReal & real & aerial & 1764 \\ 
    TargetSynth & synth & aerial & 720 \\ 
    \bottomrule
  \end{tabular}
  \label{tab:datasets}
\end{table}

The datasets we have collected provide a basis for the general boat detection from aerial images. On the other hand, three newly acquired datasets, namely MARUSBoats, TargetReal and TargetSynth, specifically target the boat model used in the competition. Consequently, our training data covers a wide range of models and conditions and is representative of the task at the competition.



\subsection{Template preprocessing}


Prior to launching an autonomous search mission, we receive visual data of a missing or unauthorized vessel, referred to as the target boat. The visual data consists of two images of the target boat, which we use as templates for the identification module. We segment the target boat using the Segment Anything Model (SAM) \cite{sam}, removing the background and non-relevant objects. We opt for SAM in this preprocessing step due to its state-of-the-art zero-shot segmentation capabilities, eliminating the need for fine-tuning.

\begin{figure}[htb]
    \centering
    \includegraphics[width=\columnwidth]{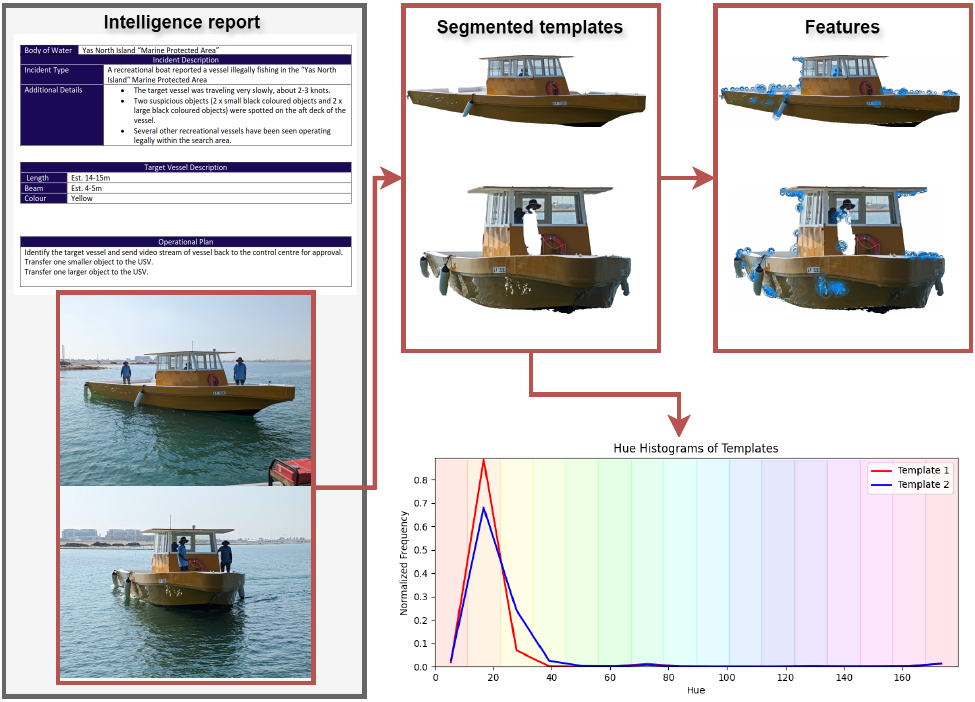}
    \caption{Template preparation process: Starting with two images from an intelligence report (left), the target boat is segmented to remove background and non-relevant elements. ORB features and hue histograms (right) are then calculated for the processed templates.}
    \label{fig:template}
\end{figure}

On segmented templates $T_1$ and $T_2$ we compute Oriented FAST and rotated BRIEF (ORB)\cite{Rublee2011} features. The features are later used in boat identification module. To utilize color information, we also calculate a hue histogram $H_k$ of the segmented template in the HSV (Hue, Saturation, Value) color space. The motivation behind hue histogram is twofold: 1) it eliminates the need for manually tuning threshold boundaries for each color or target, as the histogram automatically captures the appropriate distributions, and 2) it improves color discrimination under variable lighting conditions which is common in aerial outdoor applications. The complete template preparation process is illustrated in Fig. \ref{fig:template}.

\subsection{Target identification}

The maritime surveillance mission is fully autonomous and consists of multiple operational states, and we focus on the search state, where the UAV autonomously flies over a large open-sea area, detecting all boats and identifying the target vessel among them. When the search state is activated, the target identification module is initialized.

Upon receiving an RGB image $I \in \mathbb{R}^{H \times W \times 3}$, we employ the YOLOv8s model for object detection. YOLOv8 is a state-of-the-art model that balances accuracy and speed, making it ideal for real-time aerial operations on edge devices. Given the computational constraints of aerial platforms, we utilize the small variant, YOLOv8s, running on CPU, to ensure efficient inference. The model outputs a set of detections:

\begin{equation}
\begin{aligned}
\mathcal{D} &= \text{YOLOv8s}(I) \\ 
&= \Bigl\{ b_i = \bigl(k_i, x_c^{(i)}, y_c^{(i)}, w^{(i)}, h^{(i)}, s_i \bigr) 
\Bigr\}_{i=1}^{N_{\text{det}}}.
\end{aligned}
\end{equation}

where each detection \( b_i \) consists of:

\begin{itemize}
    \item $k_i$: Predicted class label, where $k_i \in \{0,1,\dots,K-1\}$.
    \item $(x_c^{(i)}, y_c^{(i)})$: Normalized center coordinates of the bounding box.
    \item $(w^{(i)}, h^{(i)})$: Normalized width and height of the bounding box.
    \item $s_i$: Confidence score, where $s_i \in [0,1]$.
\end{itemize}

To filter out excessively small or large detections, we apply area-based thresholding. The normalized area of each bounding box is computed as $\alpha_i = w^{(i)} h^{(i)}$, and detections outside the range

\begin{equation}
\alpha_{\min} \leq \alpha_i \leq \alpha_{\max}
\end{equation}

are discarded. Since the SAM model used for templates is too computationally expensive for real-time onboard execution, we employ a color-masking strategy to remove background from the images. Specifically, we remove blue pixels representing the sea and white pixels because their hue value is not unique when mapping from RGB to HSV color space. To achieve this, we define a binary mask function $M_{\text{bw}}(x, y)$ over each cropped bounding box $\mathbf{b}_i$:

\begin{equation}
M_{\text{bw}}(x, y) =
\begin{cases}
1, & \text{if } (x, y) \text{ is neither blue nor white}, \\
0, & \text{otherwise}.
\end{cases}
\end{equation}

We apply the mask to $b_i$. If too many pixels are removed, exceeding threshold $p_{max}$, the detection is discarded due to insufficient remaining information for identification. This step is applied unless the intelligence report specifies that blue or white boats should be considered, in which case it is disabled.

Each detection, i.e., bounding box, is a candidate for the target boat and undergoes an identification. After preprocessing, we validate each candidate compare its histogram similarity and feature matches with reference templates $T_1$ and $T_2$. For each detection $\mathbf{b}_i$, we extract ORB keypoints and compute descriptors. Given a template descriptor set $\mathcal{D}_{\text{template}}$, we use brute-force matching with Hamming distance and cross-checking to match the features. A match is considered valid if the distance is below the threshold $d_{max}$.

For further decisions we use the percentage $p_{\text{m}}$ of valid matches relative to the detected image keypoints. If set of valid matches falls below a predefined threshold, the detection is rejected. For each template, we obtain $p_{\text{m,1}}$ and $p_{\text{m,2}}$, representing the match percentages for $T_1$ and $T_2$, respectively.

Since a strong feature match with both templates is preferred, we classify detections based on their match percentage scores \( p_{\text{m,1}} \) and \( p_{\text{m,2}} \). A detection is considered a strong candidate if it has a high percentage of matched features for both templates. If one template yields a strong match while the other is moderate, the detection is considered an acceptable candidate.

Second part of identification is hue histogram distance $d_i$ between the candidate and the target templates. The histogram distance $d_i$ between the detected bounding box $\mathbf{b}_i$ and the reference histogram templates is computed using the Bhattacharyya distance:

\begin{equation}
d_{i, k} = D_{\text{Bhatt}}(H_{\text{i}}, H_k),
\end{equation}

where \( H_{\text{i}} \) is the normalized hue histogram of the detected image, and \( H_k \) represents the histograms of the reference templates, with \( k \in \{1,2\} \) corresponding to the two templates. To enforce stricter evaluation criteria, we use \( \max(d_{\text{i}, 1}, d_{\text{i}, 2}) \) as the final decision parameter.

\begin{figure}[htb]
    \centering
    \includegraphics[width=\columnwidth]{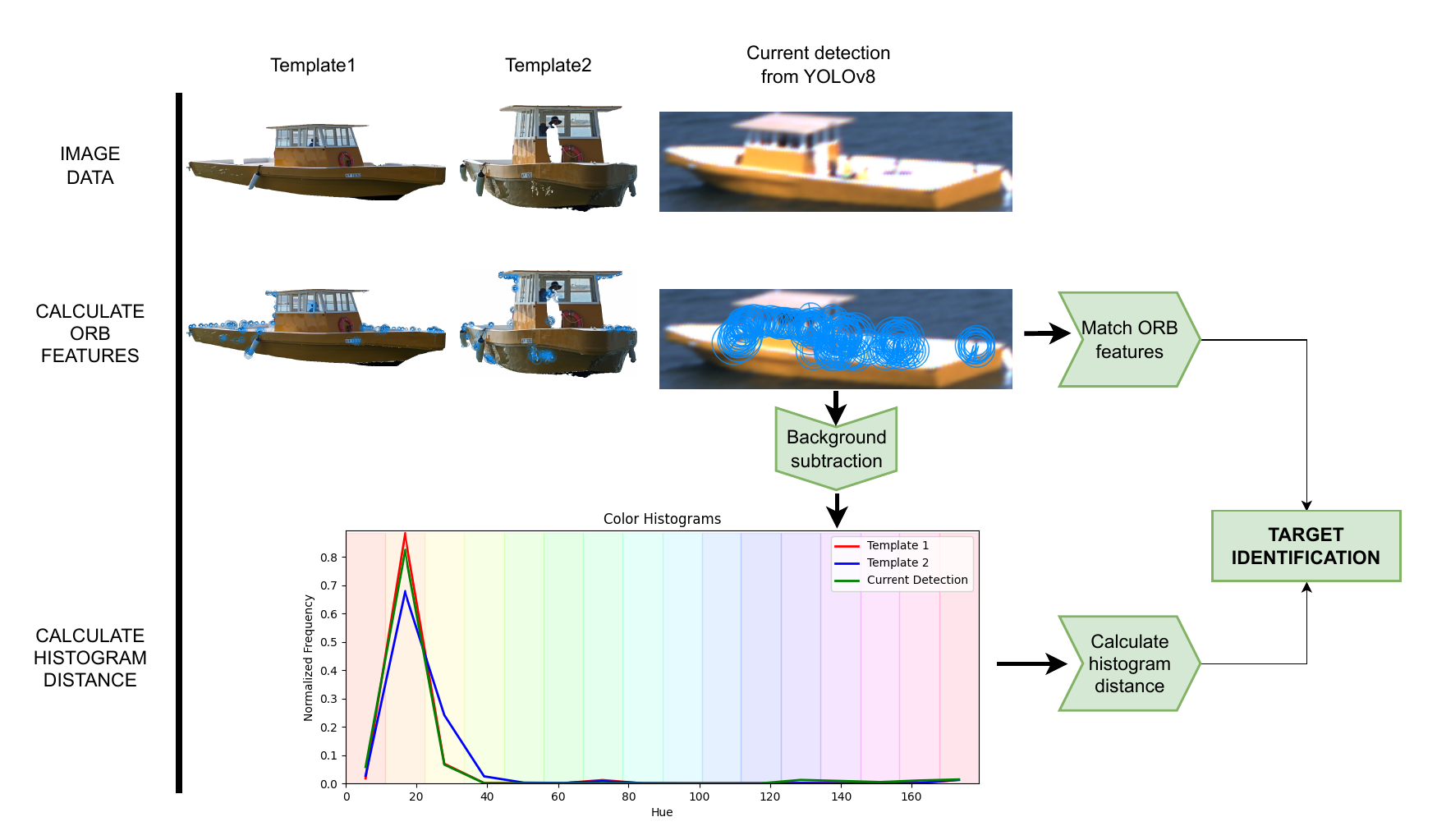}
    \caption{Target identification pipeline. The process involves performing background subtraction, extracting ORB features and computing hue histogram on a detected object \( b_i \). The detection is compared with reference templates \( T_1 \) and \( T_2 \), and a final decision is made based on feature matching and histogram distance.}
    \label{fig:target_identification}
\end{figure}

Fig. \ref{fig:target_identification} illustrates the complete target identification pipeline applied to a single detection $b_i$. The final decision on target identification is based on both the histogram distance \( d_{\text{hist}} \) and the feature match scores. The identifications is determined as follows:

\begin{itemize}
    \item If \( d_{\text{hist}} < d_{\text{certain}} \), the detection is classified as a \textbf{target}.
    \item If \( d_{\text{hist}} < d_{\text{likely}} \) and the detection is either a strong match candidate or an acceptable match, it is classified as a \textbf{possible target}.
    \item If \( d_{\text{hist}} < d_{\text{uncertain}} \) and the detection is an a strong match, it is classified as a \textbf{possible target}.
    \item If \( d_{\text{hist}} \geq d_{\text{uncertain}} \) or the detection fails to meet the minimum feature match criteria, it is classified as \textbf{not a target}.
\end{itemize}

\subsection{Position reconstruction}

To calculate the position of the target in the LoRa reference frame, in which the UAV is flying and which is shared with the USV, we calculate the direction vector $\mathbf{v}_{cam}^{vessel} = [x_{bb}, y_{bb}, f]$ that points from the camera center to the detected object with image coordinates $(x_{bb}, y_{bb})$, where $f$ is the focal length of the camera that is calibrated beforehand, and then transform the normalized direction vector into the LoRa coordinate frame:

\begin{equation}
    \mathbf{v}_{lora}^{vessel} = \mathbf{R}_{lora}^{uav} \mathbf{R}_{uav}^{cam} \mathbf{v}_{cam}^{vessel}.\label{eq:heading_calc}
\end{equation}

Intersecting the line defined with vector $\mathbf{v}_{lora}^{vessel}$ from the camera origin, obtained from localization of the UAV in the LoRa frame, with plane $z=h$ we can calculate the $x$ and $y$ positions of the target. Parameter $h$ denotes the empirical value of target height above sea level, with sea level being $z=0$. The position of the target in LoRa frame is sent to the surface vehicle which is tasked with interception.

\subsection{Human-in-the-loop}

We adopt human-in-the-loop in our decision process. After successful identification, the proof of identified target is sent to the command center. There a human operator either confirms or rejects the identified target boat, determining its validity. If the target is denied, our system resets its flags and initiates a new search. Conversely, if the target is confirmed, we proceed to navigate toward the identified boat for closer inspection and subsequent operations.
\section{Experimental Results}
\label{sec:experimental_results}

\subsection{Experimental setting}

To evaluate the proposed pipeline, we collected and manually annotated real-world experimental data from the MBZIRC 2023 competition, held in Abu Dhabi. The dataset details are provided in Table \ref{tab:mbzirc_dataset}. It includes two sequences captured on different days, featuring a target boat along with one or two decoy boats, captured at different distances from the UAV. The dataset includes two annotation sets: one providing labels for all boats on the sea, and another specifically for identification, distinguishing the target boat from other boats.

\begin{table}[h!]
    \centering
    \caption{MBZIRC2023 Abu Dhabi Boat dataset. Average object size is computed as the percentage of pixels occupied by annotated boats relative to total image pixels.}
    \label{tab:mbzirc_dataset}
    \begin{tabular}{lccc}
        \toprule
        \textbf{Dataset} & \textbf{Number of Images} & \textbf{Boats} & \textbf{Avg. object size} \\
        \midrule
        Dataset A & 513 & 2 & 0.65\%\\
        Dataset B & 996 & 3 & 0.37\%\\
        \bottomrule
    \end{tabular}
\end{table}

\subsection{Testing in Abu Dhabi}

\begin{figure}[htb]
    \centering
    \includegraphics[width=\columnwidth]{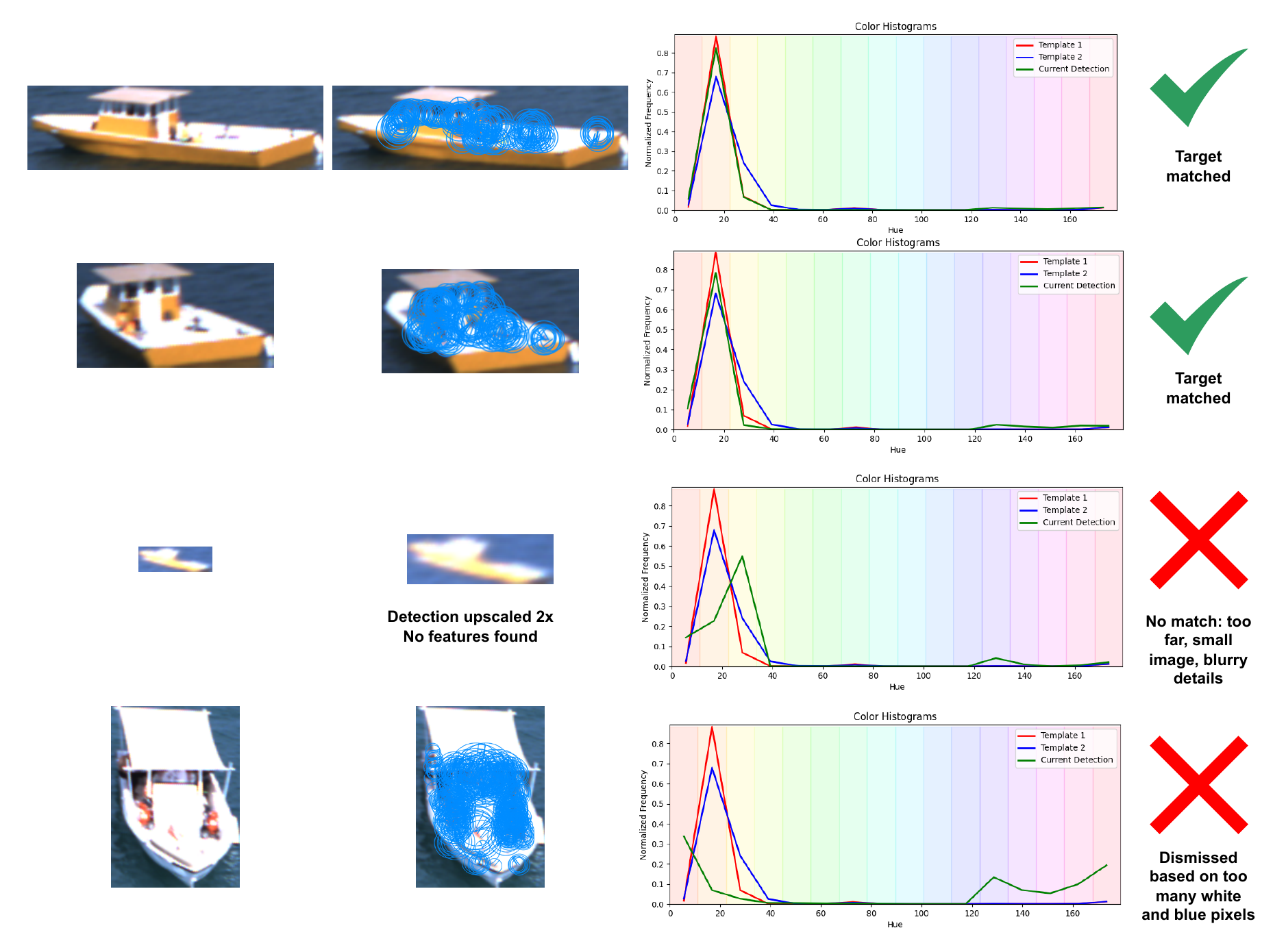}
    \caption{Examples of target identification outcomes. The first column shows candidate detections, followed by extracted ORB features, histogram comparisons, and the final identification decision. The first two rows illustrate successful identifications from different perspectives, while the third row shows a failure due to insufficient resolution. The last row depicts a correctly rejected non-target vessel.}
    \label{fig:examples}
\end{figure}

To quantitatively evaluate our system, we analyze data from the MBZIRC2023 dataset. Fig. \ref{fig:examples} illustrates representative scenarios encountered during target identification. For each example, we display the detected features and corresponding histograms, while the templates are always the same and their associated features and histograms are shown in Fig. \ref{fig:template}.  

First row shows an example where we have a good visual representation of the target, resulting in a strong feature match and aligned histograms, and the target has been successfully identified. The second example also demonstrates successful identification, but from a completely different perspective—capturing the boat from the rear. This confirms the system's ability to handle viewpoint variations, including significant rotations. The third example shows a challenging scenario where the detected boat is too far away, resulting in a low resolution detection with insufficient pixel information. Even after upscaling, feature extraction was unsuccessful, and due to highly illuminated object there was a shift in the histogram. This shows that the system reaches its limits when the target is too far away, as the visual cues are less informative. The last example illustrates a rejected detection. In this case, the detected boat is not the target as the feature matching fails and the histogram distance is too large.

\begin{figure}[htb]
    \centering
    \includegraphics[width=\columnwidth]{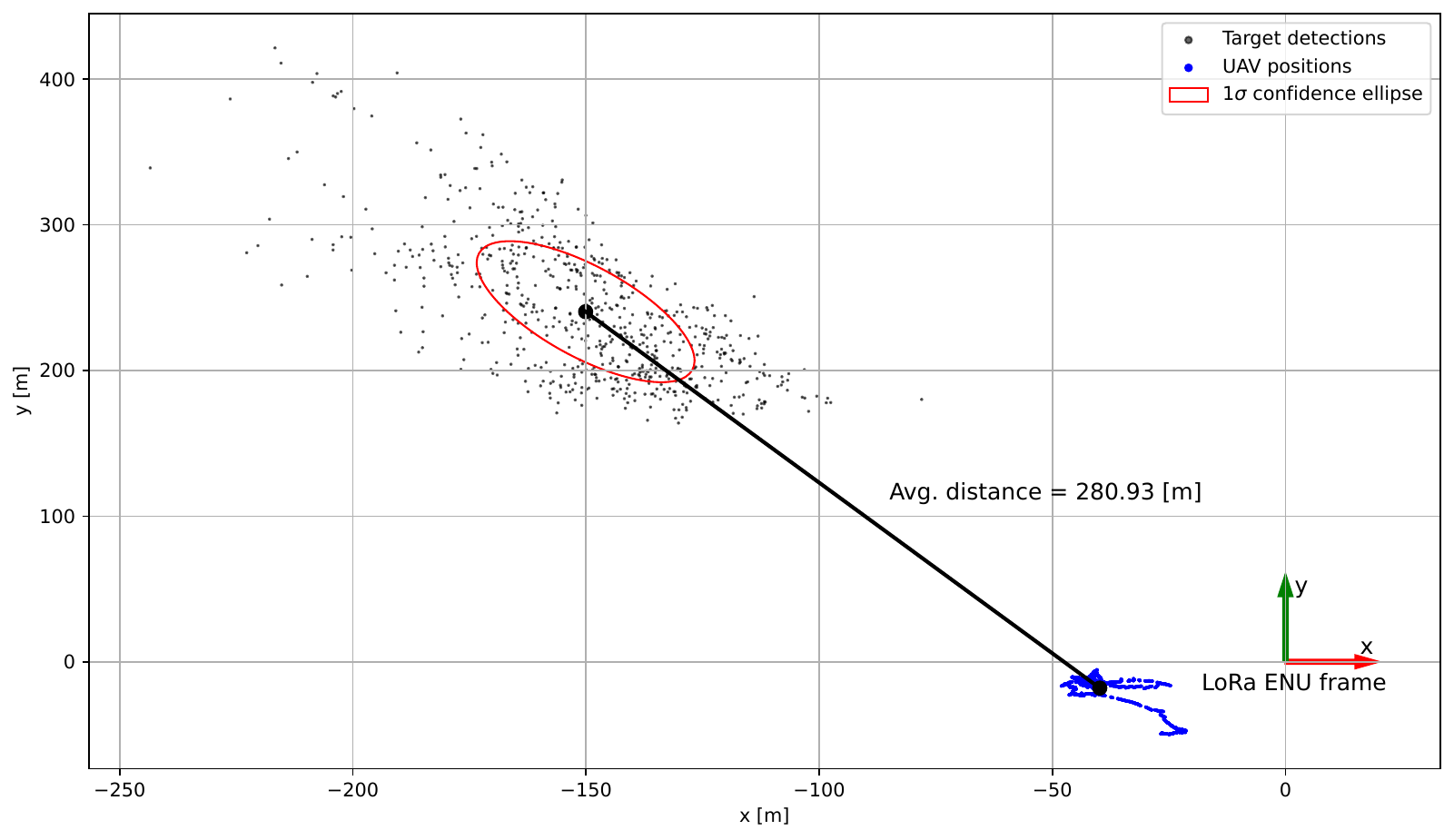}
    \caption{UAV positions and reconstructed positions of the target in the LoRa ENU coordinate frame, data from Dataset B. Red ellipse denotes $1\sigma$ confidence area for the position of the target.}
    \label{fig:position_reconstructed}
\end{figure}

Fig. \ref{fig:position_reconstructed} shows the positions of the UAV and the reconstructed positions of the target, with $1\sigma$ confidence ellipse. 

While the Fig. \ref{fig:position_reconstructed} shows the average distance between UAV and the target to be 280 meters, the identification module can reliably detect MBZIRC target boats up to 500 meters from the UAV. The scattering of the detections arises from sensitivity of distance to image resolution (1px change in the center of the box translates to tens of meters change in the estimate), but is also heavily influenced by imperfect synchronization between detection/identification and the UAV attitude.

\subsection{Ablation study}

\textbf{Boat-to-boat vs. aerial perspective.} To demonstrate importance of perspective represented by the training data we compare two identical object detector models trained on data with different perspectives. The first detector is trained only on boat-to-boat perspective, and second detector is trained on collection of public and our datasets with both boat-to-boat and aerial perspective as detailed in Table \ref{tab:datasets}. Both object detection models are evaluated on MBZIRC2023 Abu Dhabi datasets (\ref{tab:mbzirc_dataset}) and results are shown in Table \ref{tab:perspective}.


\begin{table}[h!]
    \centering
    \caption{An evaluation of detection model performance based on training perspectives: boat-to-boat (b2b) and aerial (a). Models were tested on the MBZIRC2023 Abu Dhabi dataset to analyze the influence of perspective on detection accuracy.}
    \setlength{\tabcolsep}{1.2pt} 
    \label{tab:perspective}
    \begin{tabular}{lccccc}
        \toprule
        \textbf{Model} & \textbf{Perspective} & \multicolumn{2}{c}{\textbf{Dataset A}} & \multicolumn{2}{c}{\textbf{Dataset B}} \\
        \cmidrule(lr){3-4} \cmidrule(lr){5-6}
        & &  mAP50 & mAP50:95 & mAP50 & mAP50:95 \\
        \midrule
        Datasense & b2b  & 0.867 & 0.517 &  0.946 & 0.558\\
        Ours & b2b+a & \textbf{0.929} & \textbf{0.773} & \textbf{0.992} & \textbf{0.817} \\
        \bottomrule
    \end{tabular}
\end{table}

The model trained solely on a boat-to-boat perspective demonstrates lower mean average precision (mAP) in both Dataset A and Dataset B, even with a forward-facing aerial perspective, which is the most similar aerial angle to the boat-to-boat view. Minimal differences were therefore expected. However, the results indicate that incorporating aerial perspectives provides more diverse training data, significantly improving the model's performance.

In robotics, precise localization is crucial, as our target position reconstruction heavily depends on the accuracy of detections. To assess this, we compared mAP across a range of IoU thresholds (mAP50:95). The results reveal an even greater improvement in performance when aerial images are included in the training dataset.

\begin{figure}[htb]
  \begin{subfigure}{0.49\columnwidth}
    \includegraphics[width=\columnwidth]{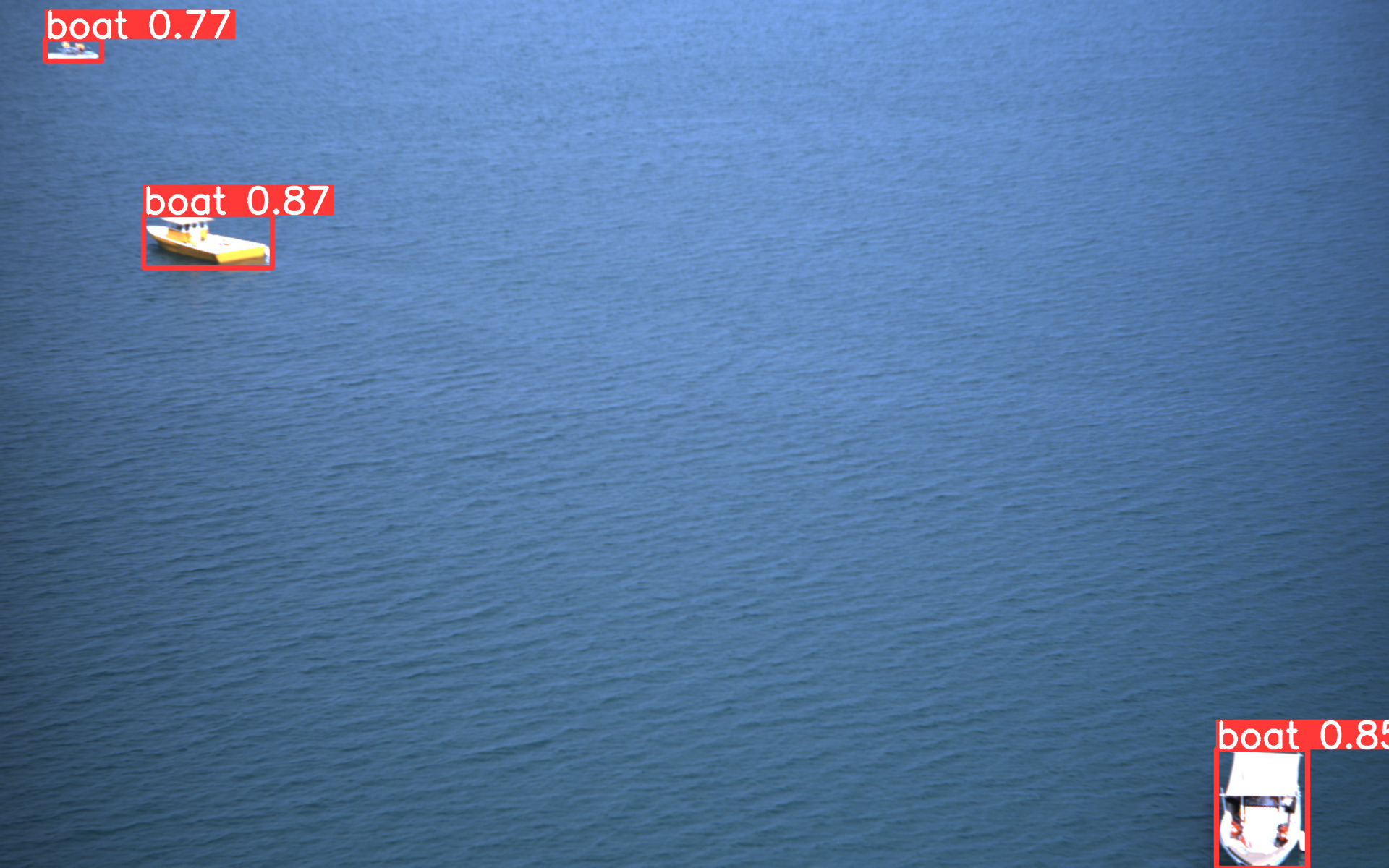}
    \caption{Model trained on b2b+a}
    \label{fig:cover-a}
  \end{subfigure}%
  \hfill
  \begin{subfigure}{0.49\columnwidth}
    \includegraphics[width=\columnwidth]{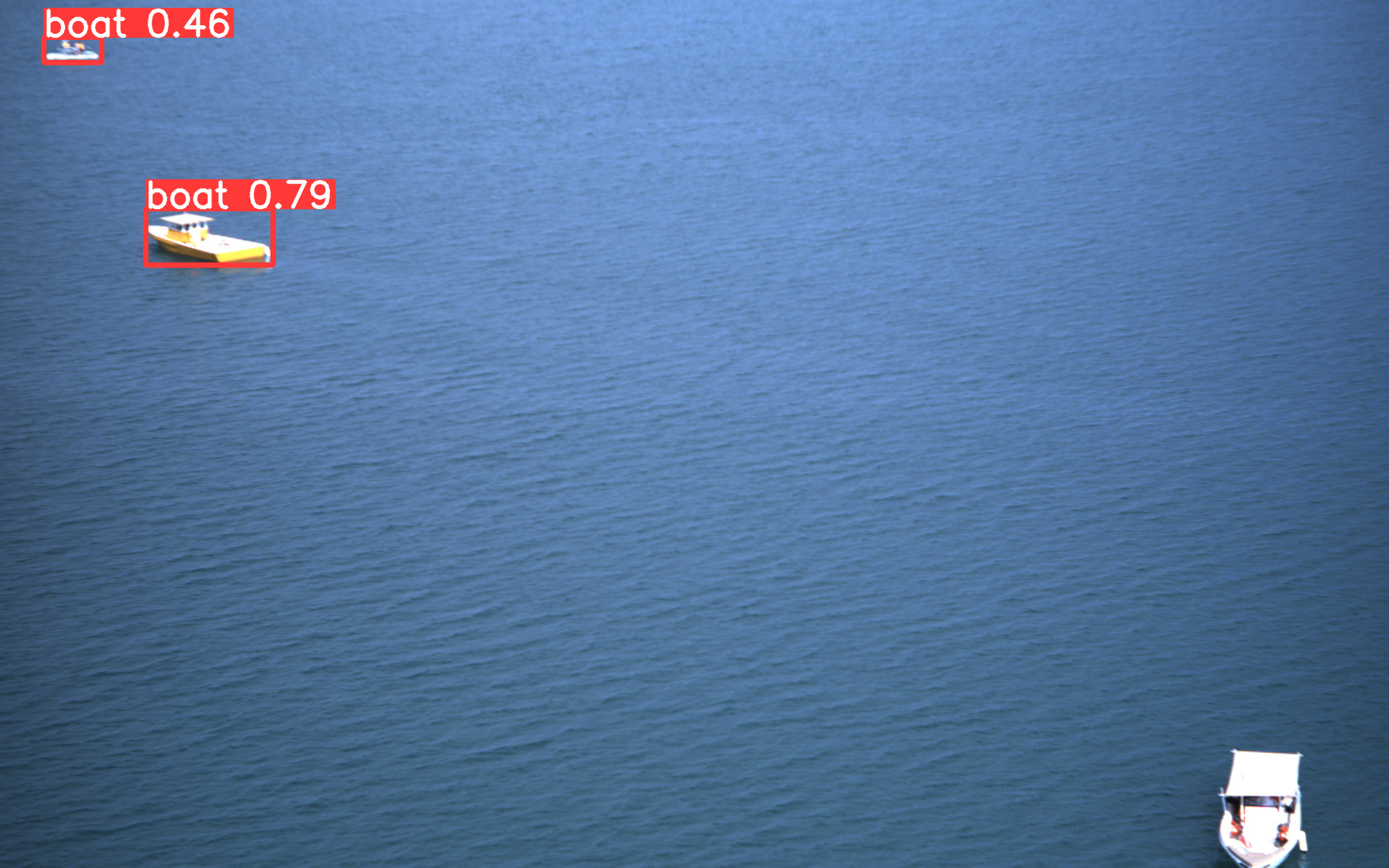}
    \caption{Model trained on b2b only}
    \label{fig:cover-b}
  \end{subfigure}
  \caption{Qualitative comparison of the influence of training data perspectives on detection model performance. The model trained without aerial perspective (b) misses one detection and demonstrates lower confidence in detections. }
  \label{fig:both-perspectives}
\end{figure}

Fig. \ref{fig:both-perspectives} shows a qualitative comparison where the boat-to-boat model exhibits lower confidence and occasionally misses boats in aerial views, while the combined model demonstrates more consistent detection.

\textbf{Template matching vs. object detector with \textit{target boat} class.} We compare two approaches for identification of the target boat in open-sea scenarios where multiple boats are present. The first approach utilizes our identification module, which combines a ship detector with feature and histogram matching based on predefined templates. The second approach involves training an object detector with two classes: \textit{boat} and \textit{target boat}. To achieve this, we structure the training data as follows: synthetic and real-world images of the target boat are labeled as \textit{target boat}, while all other boats are assigned the \textit{boat} class. 

\begin{figure}[htb]
    \centering
    \begin{subfigure}{0.498\columnwidth}
        \centering
        \includegraphics[width=\columnwidth]{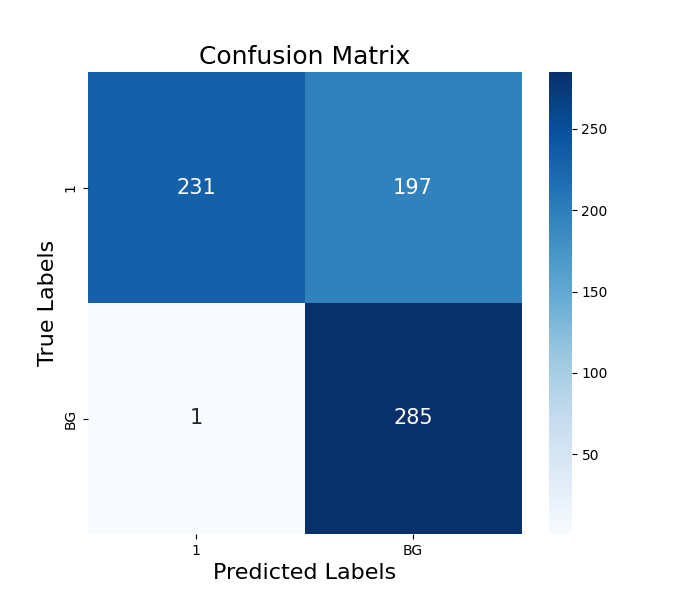}
        \label{fig:sub1}
        \caption{Dataset A - ID}
    \end{subfigure}%
    \hfill
    \begin{subfigure}{0.498\columnwidth}
        \centering
        \includegraphics[width=\columnwidth]{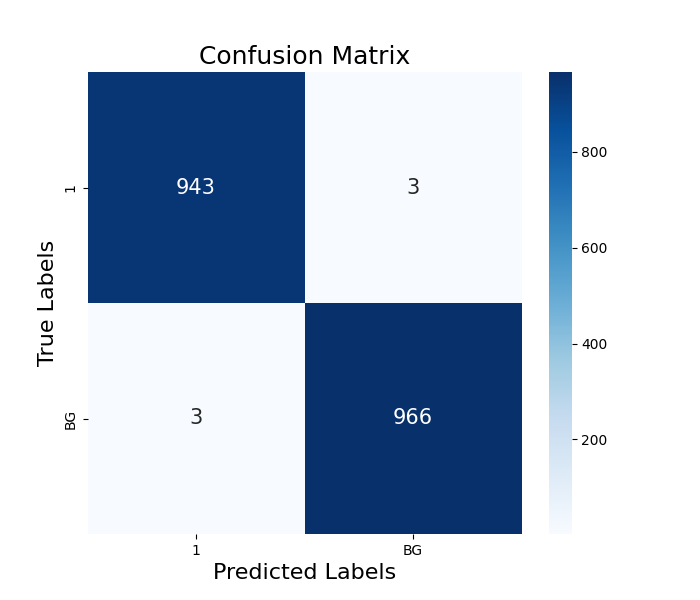}
        \label{fig:sub2}
        \caption{Dataset B - ID}
    \end{subfigure}
    
    \begin{subfigure}{0.498\columnwidth}
        \centering
        \includegraphics[width=\columnwidth]{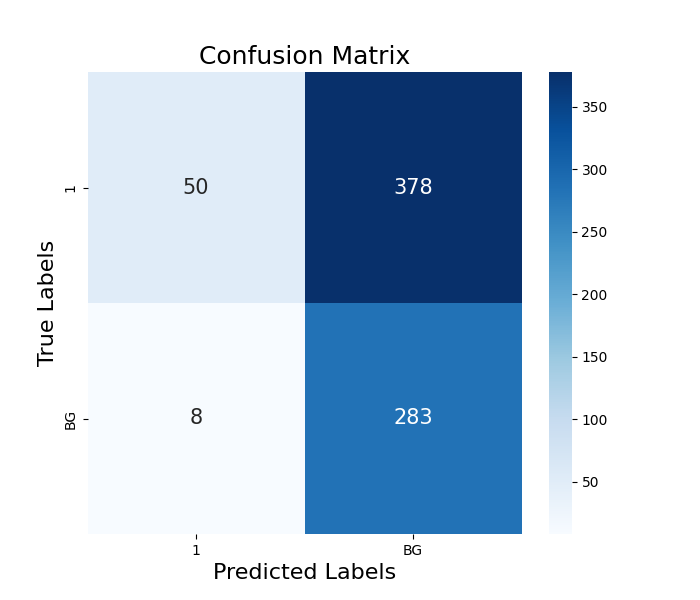}
        \caption{Dataset A - YOLO}
        \label{fig:sub3}
    \end{subfigure}%
    \hfill
    \begin{subfigure}{0.498\columnwidth}
        \centering
        \includegraphics[width=\columnwidth]{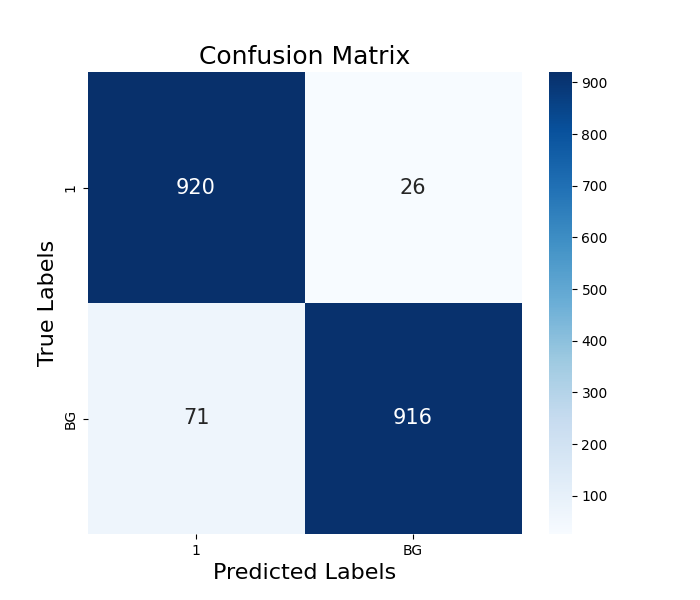}
        \caption{Dataset B - YOLO}
        \label{fig:sub4}
    \end{subfigure}
    \caption{Comparison of target identification approaches: the proposed identification (ID) module vs YOLO with the \textit{target boat} class.}
    \label{fig:confusion-matrix}
\end{figure}

\begin{figure}[htb]
  \centering
  \begin{subfigure}{0.498\columnwidth}
    \includegraphics[width=\columnwidth]{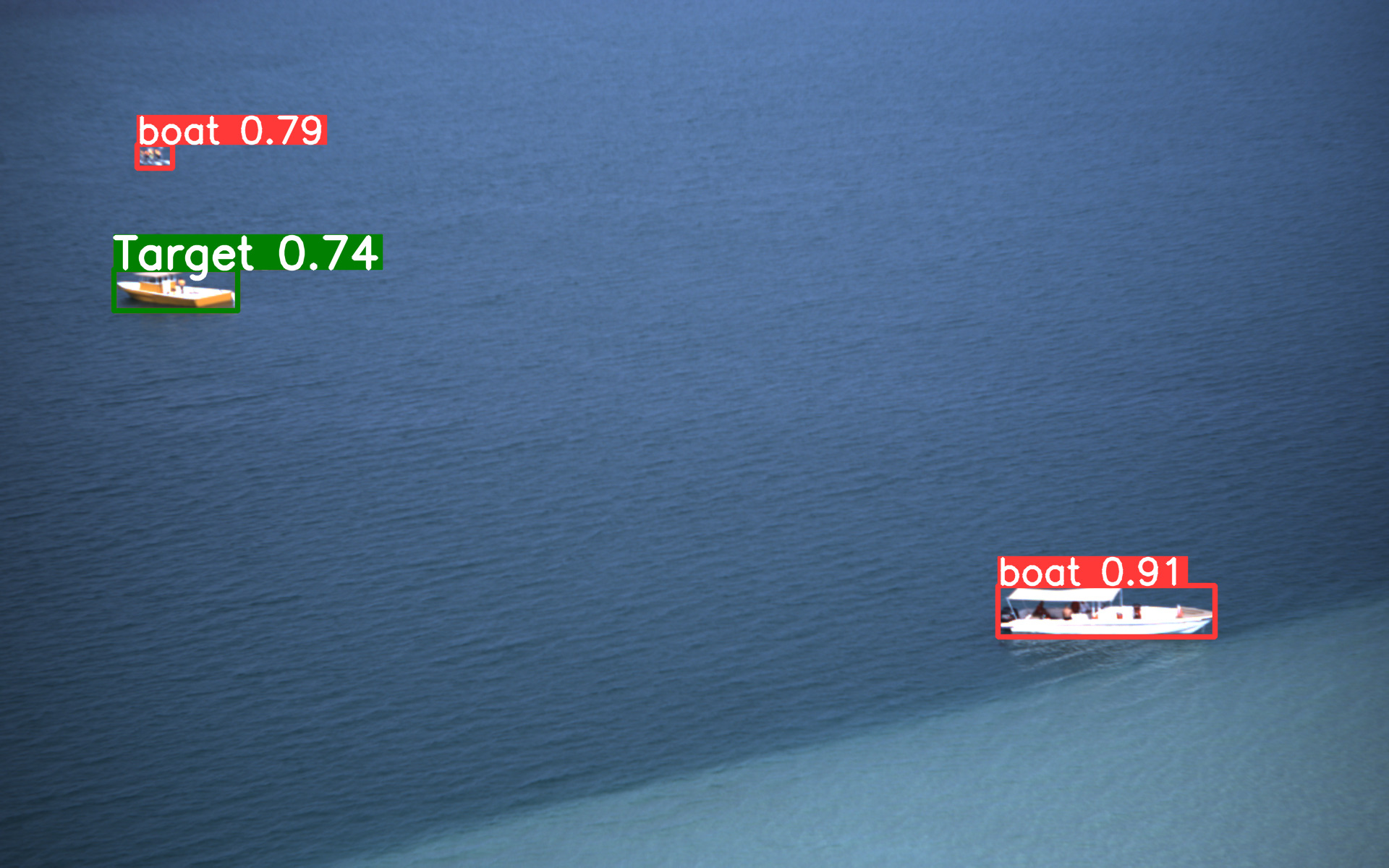}
    \label{fig:aerial-view}
  \end{subfigure}%
  \hfill
  \begin{subfigure}{0.498\columnwidth}
    \includegraphics[width=\columnwidth]{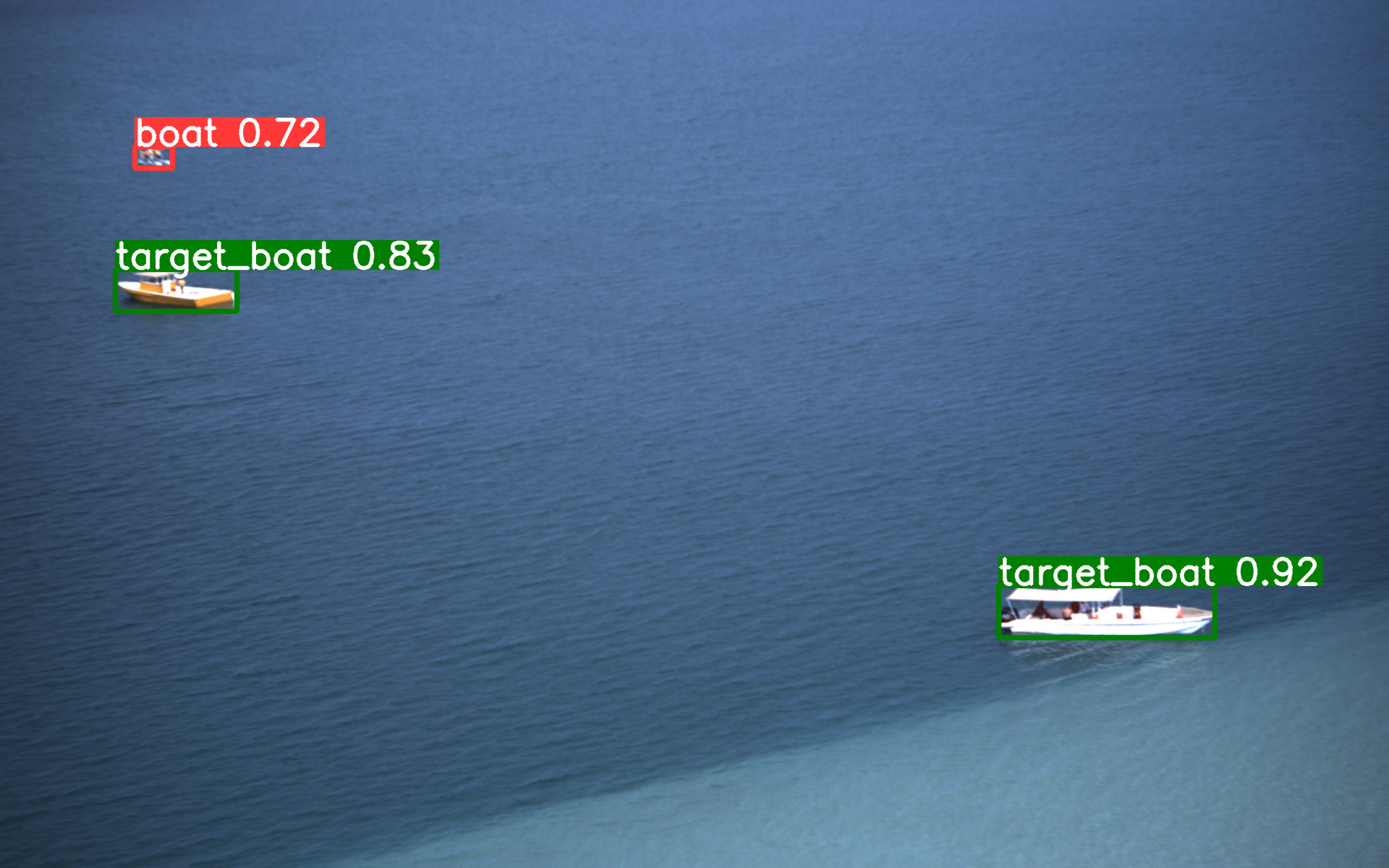}
    \label{fig:boat-view}
  \end{subfigure}
  \begin{subfigure}{0.498\columnwidth}
    \includegraphics[width=\columnwidth]{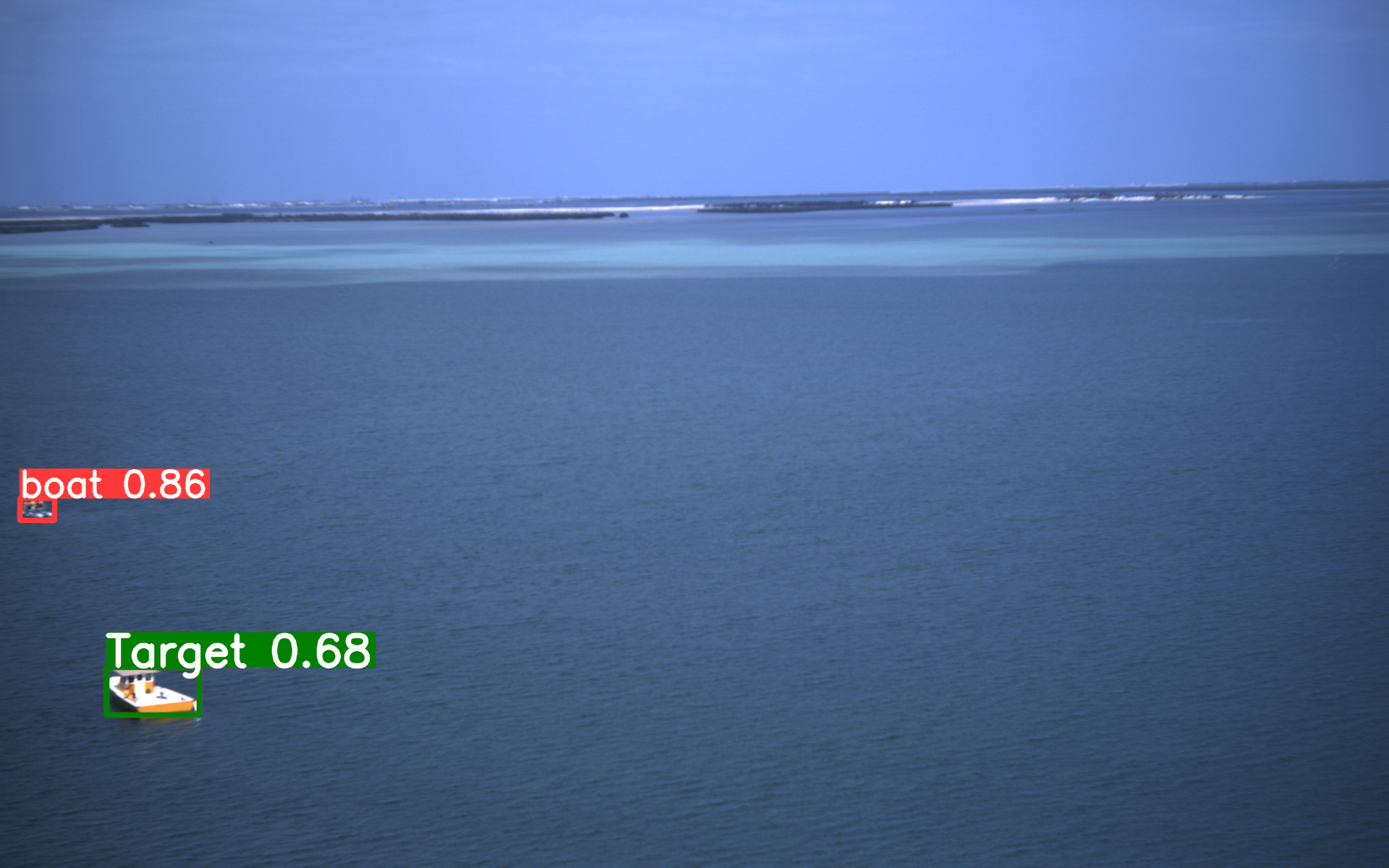}
    \caption{ID}
    \label{fig:third-perspective}
  \end{subfigure}%
  \hfill
  \begin{subfigure}{0.498\columnwidth}
    \includegraphics[width=\columnwidth]{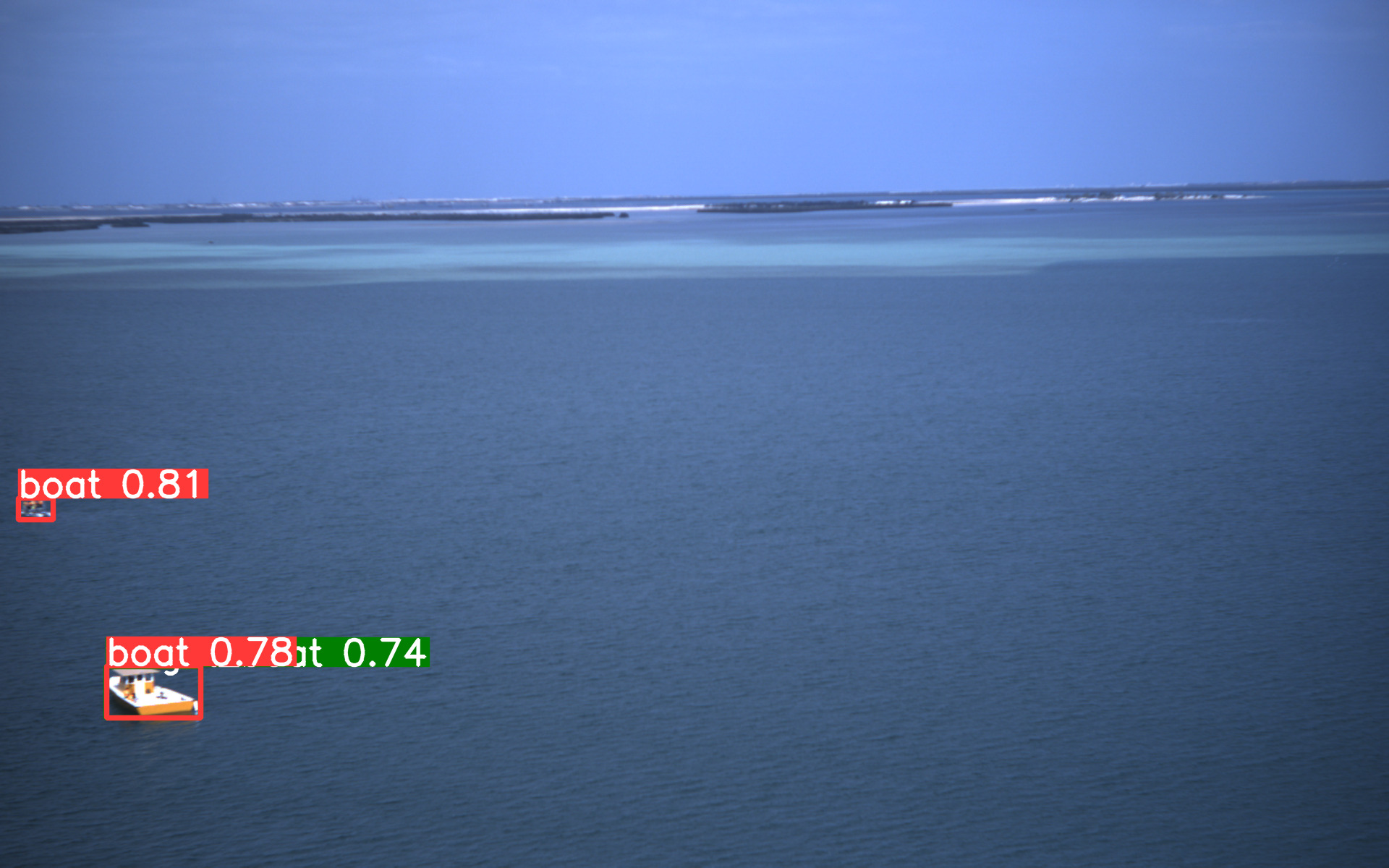}
    \caption{YOLO}
    \label{fig:fourth-perspective}
  \end{subfigure}
  \caption{Snapshot of identification of the target with two analyzed approaches: the proposed identification (ID) module vs YOLO with the \textit{target boat} class. In the first row, YOLO incorrectly identifies the wrong boat as the target, while in the second row, it fails to distinguish between a boat and the target boat. Our ID module performs reliably.}
  \label{fig:id-vs-yolo}
\end{figure}

We evaluate both approaches on Dataset A and Dataset B from the MBZIRC dataset. The results in terms of confusion matrix are presented in Fig. \ref{fig:confusion-matrix}, while Fig. \ref{fig:id-vs-yolo} shows an snapshot of the mission where two detectors behave differently. In this identification task, we define true negatives as cases where a boat was present, but the approach correctly did not classify it as a target boat. Our approach achieves a precision of $99.57\%$ and a recall of $53.97\%$ on Dataset A, while the two-class detector achieves a precision of $86.21\%$ and a recall of $11.68\%$. Similarly, our approach achieves a precision of $99.68\%$ and a recall of $99.68\%$ on Dataset B, while the two-class detector achieves a precision of $92.84\%$ and a recall of $97.25\%$. The results show that our proposed method of identification by template matching achieves higher precision and recall values across both datasets.

The results are somewhat unexpected, as the two-class detector in this context acts as an Oracle, an idealized model with perfect information, i.e. a fully labeled dataset of the target boat. However, in real-world it is not possible to have a pre-existing dataset of the target, along with the time required for labeling and training the model. Thus, this approach is not realistically achievable with available resources. More importantly, our method outperforms it in terms of identification performance.

To explain why our approach performs better, we hypothesize that in aerial imagery, the small size of the target boat, resulting from the large distance between the UAV and objects at sea, makes it difficult for a neural network to detect discriminative features between the \textit{boat} and \textit{target boat} classes. While the network performs well in favorable conditions, it frequently produces false positive target identifications, which are highly undesirable for the problem addressed in this paper. In contrast, the traditional approach of feature analysis and hue histogram comparison, has proven to be a more reliable solution.





\section{Conclusion}
\label{sec:conclusion}

In this paper, we have presented the identification module for aerial surveillance, recognition and localization of a known target vessel in maritime environment. Due to constraints on power consumption and weight, small version of YOLOV8 detector network is executed on the onboard CPU of the UAV for general ship detection, while identification is performed using traditional computer vision techniques via feature and color matching. To validate performance of the identification module, we have presented an ablation study, analyzing the influence on perspective in the training dataset and comparing the proposed identification module with an Oracle trained on the exact images of the target. Localization is performed using simple geometric principles, and was shown to be effective for target over 300 meters away. The performance of the target identification and localization module, as shown in this paper, was a key factor in enabling the UNIZG-FER team to achieve first place at the MBZIRC 2023 Grand Maritime Challenge.

\section*{ACKNOWLEDGMENT}
\small{This work has been supported by the European Union’s Horizon Europe research program Widening participation and spreading excellence, through project
Strengthening Research and Innovation Excellence in Autonomous Aerial
Systems (AeroSTREAM) - Grant agreement ID: 101071270}. Authors would like to thank UNIZG-FER MBZIRC team members for their work on localization and control of the surveillance UAV.


\bibliographystyle{ieeetr}
\bibliography{bibliography/Mendeley}

\end{document}